\documentclass[letterpaper, 10 pt, conference]{ieeeconf}  

\IEEEoverridecommandlockouts                              

\title{\LARGE \bf
Distilled Visual and Robot Kinematics Embeddings for Metric Depth Estimation in Monocular Scene Reconstruction
}

\author{
Ruofeng Wei$^{1,\dag}$, Bin Li$^{2,\dag}$, Hangjie Mo$^{1}$, Fangxun Zhong$^{2}$, Yonghao Long$^{3}$,  \\Qi Dou$^{3},\textit{Member, IEEE}$, Yun-Hui Liu$^{2},\textit{Fellow, IEEE}$ and Dong Sun$^{1,*},\textit{Fellow, IEEE}$
\thanks{*This work was supported by the Research Grants Council of the Hong Kong Special Administrative Region, China (Ref. No. T42-409/18-R and 11211421), and a grant from City University of Hong Kong (Ref. No. 7005447). (Corresponding author: Dong Sun).}
\thanks{$^{1}$Ruofeng Wei and Hangjie Mo are with the Department of Biomedical Engineering, City University of Hong Kong, Hong Kong, China (E-mail: ruofenwei2-c@my.cityu.edu.hk; hangjiemo2-c@my.cityu.edu.hk).}%
\thanks{$^{2}$Bin Li, Fangxun Zhong, and Yunhui Liu are with the Department of Mechanical and Automation Engineering, Chinese University of Hong Kong, Hong Kong, China (Email: bli@mae.cuhk.edu.hk; fxzhong@cuhk.edu.hk; yhliu@mae.cuhk.edu.hk).}%
\thanks{$^{3}$Yonghao Long, Qi Dou are with the Department of Computer Science and Engineering, and T Stone Robotics Institute, The Chinese University of Hong Kong, Hong Kong, China (Email: yhlong@cse.cuhk.edu.hk; qidou@cuhk.edu.hk).}%
\thanks{$^{1}$Dong Sun is with the Department of Biomedical Engineering, City University of Hong Kong, Hong Kong, China (e-mail: medsun@cityu.edu.hk).}
\thanks{The first two authors contributed equally.}%
}

\usepackage[utf8]{inputenc}
\usepackage[T1]{fontenc}
\usepackage{multirow}
\usepackage{graphicx}
\usepackage{svg}
\usepackage{amsmath}

\usepackage{amsfonts,amssymb}
\usepackage{xcolor}
\usepackage{bm}

\usepackage{booktabs}
\usepackage{colortbl}

\usepackage{gensymb}

\usepackage[ruled,linesnumbered]{algorithm2e}

\makeatletter
\renewcommand{\maketag@@@}[1]{\hbox{\m@th\normalsize\normalfont#1}}%
\makeatother

\begin{document}

\maketitle
\thispagestyle{empty}
\pagestyle{empty}

\begin{abstract}

Estimating precise metric depth and scene reconstruction from monocular endoscopy is a fundamental task for surgical navigation in robotic surgery. 
However, traditional stereo matching adopts binocular images to perceive the depth information, which is difficult to transfer to the soft robotics-based surgical systems due to the use of monocular endoscopy.
In this paper, we present a novel framework that combines robot kinematics and monocular endoscope images with deep unsupervised learning into a single network for metric depth estimation and then achieve 3D reconstruction of complex anatomy. 
Specifically, we first obtain the relative depth maps of surgical scenes by leveraging a brightness-aware monocular depth estimation method. Then, the corresponding endoscope poses are computed based on non-linear optimization of geometric and photometric reprojection residuals.
Afterwards, we develop a Depth-driven Sliding Optimization (DDSO) algorithm to extract the scaling coefficient from kinematics and calculated poses offline. By coupling the metric scale and relative depth data, we form a robust ensemble that represents the metric and consistent depth. Next, we treat the ensemble as supervisory labels to train a metric depth estimation network for surgeries (\emph{i.e.}, \emph{MetricDepthS-Net}) that distills the embeddings from the robot kinematics, endoscopic videos, and poses. With accurate metric depth estimation, we utilize a dense visual reconstruction method to recover the 3D structure of the whole surgical site. 
We have extensively evaluated the proposed framework on public SCARED and achieved comparable performance with stereo-based depth estimation methods. Our results demonstrate the feasibility of the proposed approach to recover the metric depth and 3D structure with monocular inputs.

\end{abstract}

\section{INTRODUCTION}

Transoral Robotic Surgery (TORS) has gradually become a valid treatment for tumor resection through the mouth \cite{meulemans2022transoral}.
Nowadays, researchers have developed soft robotics technology-based flexible surgical systems for TORS \cite{fang2021soft}. Different from open abdominal surgeries and robot-assisted minimally invasive surgeries (R-MIS), the endoscope utilized in TORS usually provides monocular images for surgeons. The surgical system is unable to produce 3D data for the whole surgical field due to the restricted field-of-view (FoV), which impairs the surgeon’s understanding of patient's anatomy \cite{maier2013optical}. 
Thus, the metric depth perception and the scale aware 3D surgical scene reconstruction from the monocular endoscopic sequences are highly demanded.

\begin{figure*}[htp]
    \centering
    \includegraphics[width = 0.9\hsize]{"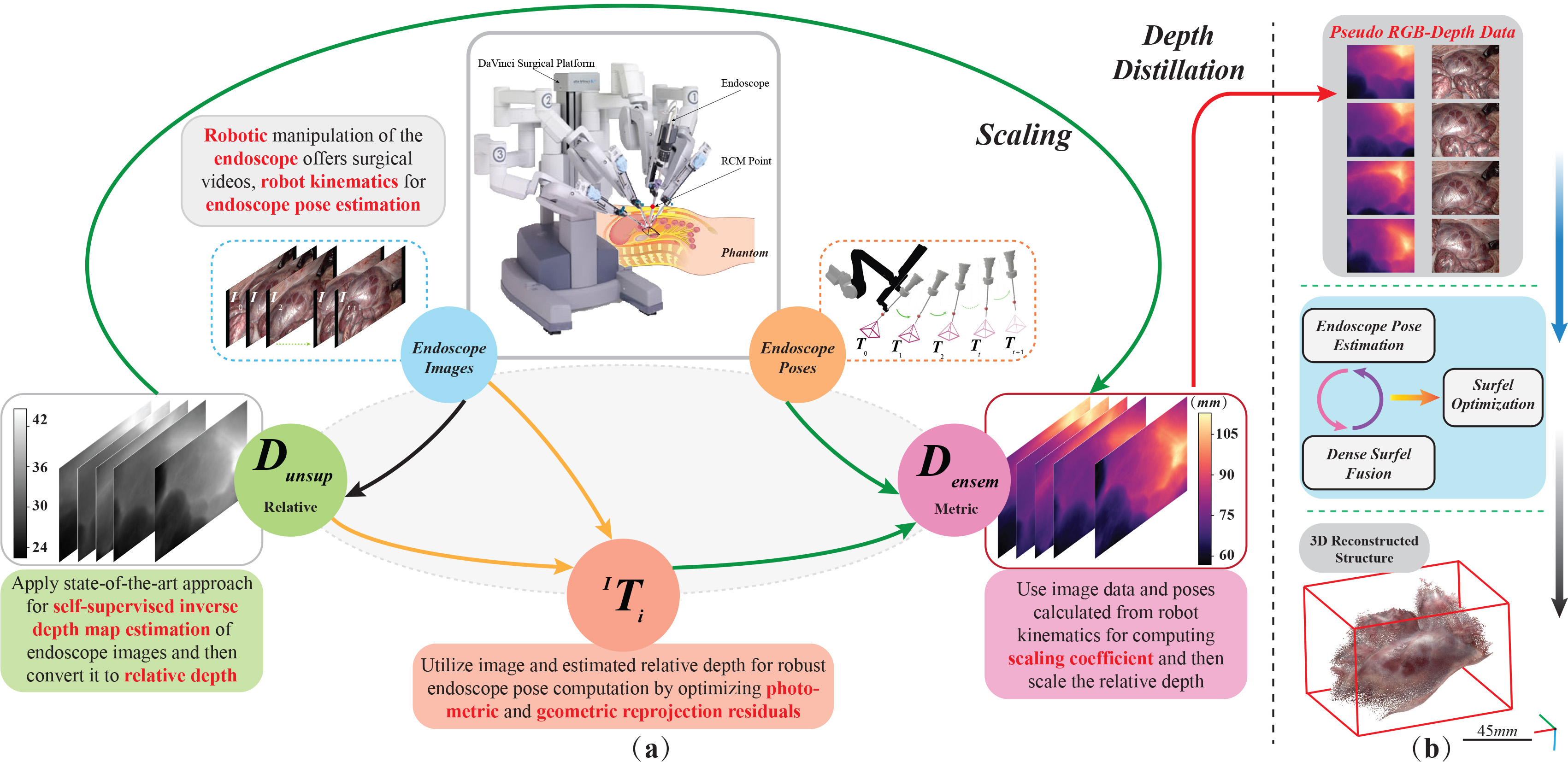"}
    \caption{Overview of the proposed framework for metric depth estimation and 3D reconstruction. (a) The depth estimation module, combining several methods to achieve accurate scale depth estimation for monocular endoscopy. (b) The process of dense visual reconstruction of tissue surfaces.}
    \label{overall_Framework}
\end{figure*}

Monocular depth estimation has attracted increasing attention in recent years. Zhou et al. \cite{zhou2017unsupervised} pioneeringly proposed a self-supervised framework which is trained by photometric reprojection loss. The photometric loss was calculated using the difference between the target frame and the novel view synthesized with estimated depth and camera motion. Since then, many extensive works have been performed by adding more blocks, such as geometric priors \cite{yang2018lego} and self-attention mechanism \cite{zhou2019unsupervised}, to improve the accuracy of depth estimation. While these methods possessed good depth estimation ability in urban scenes, they performed poorly in the textureless surgical environment \cite{shao2021self}. In medical contexts, Huang et al. \cite{huang2021self} presented a self-supervised adversarial depth estimation framework, which required stereo laparoscopic images for training.
Liu et al. \cite{liu2019dense} proposed an unsupervised monocular dense depth estimation network for functional endoscopic sinus surgeries (FESS) with sparse depth derived from Structure from Motion (SfM) \cite{schonberger2016structure}.
The brightness variation among endoscopic streams was involved for the first time to aid the depth estimation task \cite{shao2021self}.
However, these methods either adopted stereo endoscopic images for training \cite{huang2021self} or were arbitrarily scaled with respect to the real world and thus cannot be applied to TORS for monocular metric depth estimation without access to external sensors. 

Robotic systems embedded in surgical procedures enable efficient manipulation control of instruments and endoscope, yielding rich sources of information, e.g., robot kinematics. Amsterdam et al. \cite{van2022gesture} adopted the multi-modal attention mechanisms to integrate the kinematics and visual features in a surgical gesture recognition system. The work of MRG-Net \cite{long2021relational} exploits surgical videos and robot kinematics for automatic surgical gesture recognition. Colleoni et al. \cite{colleoni2020synthetic} proposed a learning-based network that employs the laparoscope images and robot kinematics to achieve surgical instrument segmentation. However, these works are related to surgical gesture recognition and instrument segmentation instead of surgical scene depth estimation. Li et al. \cite{li2021data} utilized the robotic information from Universal Robot (UR) to compute laparoscope poses and then merged the poses into a convolutional neural network (CNN) to recover metric depth. However, this method only estimates the depth of an instrument in surgical sites. In the urban environment, many works combined monocular camera and additional sensors equipped in UAVs or vehicles to accurately estimate metric depth and camera poses. Pirvu et al. \cite{pirvu2021depth} distilled the metric depth calculated from GPS trajectory information and optical flow into a single network for real-scale depth estimation. 
However, to the best of our knowledge, the metric depth estimation for the surgical scene using robot kinematics and monocular endoscopic videos has not been exploited.

In this paper, we propose a novel learning-driven architecture to achieve accurate metric depth estimation of monocular sequences and further reconstruct the dense 3D structure of tissue surfaces. Our main contributions are:

\begin{itemize}
    \item We, for the first time, propose an unsupervised learning framework for metric depth estimation of surgical scenes, which is capable of distilling the embeddings from monocular endoscopic videos and robot kinematics into a single deep network. 
    \item A new depth-driven sliding optimization (DDSO) scheme is exploited to estimate precise and smooth scaling coefficient from robot kinematics and monocular sequences, which can be used to form the visual and kinematics embeddings. 
    \item Based on the SCARED data, we conduct quantitative and qualitative experiments to evaluate the proposed framework. Our method achieves a comparable metric depth estimation performance to typical stereo-based approaches.
\end{itemize}


\section{METHODS}
The overview of our proposed framework is shown in Fig. \ref{overall_Framework}. Our approach consists of four complementary ways for metric depth estimation and dense reconstruction. Along the first path (black arrow), we estimate the inverse depth map of the monocular image in a self-supervised way and then convert it to relative depth (\emph{$D_{unsup}$}). Along the second path (yellow arrows), we compute the endoscope poses (\emph{${}^I\bm{T}_i$}) under image coordinate system through minimizing photometric and geometric reprojection residuals. Along the third path (green arrows), \emph{${}^I\bm{T}_i$} is used to calculate the scaling coefficient \emph{r} by  combining robot kinematics. Then, \emph{r} is utilized to scale \emph{$D_{unsup}$} to make it be metric. The two (\emph{r} and \emph{$D_{unsup}$}) formulate an ensemble (\emph{$D_{ensem}$}), used to distill a single network \emph{MetricDepthS-Net} for metric depth estimation of monocular surgical scene (Fig. \ref{depth_distill_network}). Along the last path (Fig. \ref{overall_Framework}(b)), we use the network \emph{MetricDepthS-Net} to generate pseudo RGB-Depth data, and then achieve a dense visual reconstruction of surgical sites.

\subsection{Inverse depth map estimation for endoscope images}

We use the Monodepth2 \cite{godard2019digging} architecture as the backbone and adopt its training process, which is a typical self-supervised depth estimation method. During the training phase, the key self-supervisory signal is calculated from novel warping-based view synthesis, which includes computed inverse depth maps, endoscope intrinsics, and camera motion. Once the per-pixel inverse depth values of the target image are estimated, we convert the pixels to relative depth map and then project it to the new image coordinate. Given the target image $I_t(\bm{p})$ and source image $I_s(\bm{p})$, the view synthesis can be written as:
\begin{align}
h(\bm{p}_{s \to t}) \sim \bm{K} \cdot {}^s\bm{T}_t \cdot \frac{1}{\emph{\text{ID}}_t(\bm{p})} \cdot \bm{K}^{-1} \cdot h(\bm{p}_t)
\end{align}
where $\bm{p}$ denotes the image pixel's 2D coordinates, $\bm{K}$ is the endoscope intrinsics, ${}^s\bm{T}_t$ denotes the rigid transformation from target view (\emph{t}) to source view (\emph{s}), which is predicted by another pose estimation net, $\emph{\text{ID}}_t(\bm{p})$ is the inverse depth in the target view and \emph{h} is the function that transforms the pixel's 2D coordinates into homogeneous coordinates. Then, we can generate the synthesized target image $I_{s \to t}(\bm{p})$ from the source image as:
\begin{align}
I_{s \to t}(\bm{p}) = I_s \left \langle \bm{p}_{s \to t}  \right \rangle
\end{align}
where $\left \langle \cdot \right\rangle$ is the sampling operator. Based on the brightness constancy assumption, the network is learned by minimizing the photometric loss between $I_t(\bm{p})$ and $I_{s \to t}(\bm{p})$.

In robotic surgery, illumination changes and non-Lambertian reflection can cause serious image brightness variation and thus break down the assumption heavily. According to \cite{shao2022self}, an appearance module AFNet is used to predict appearance flow and calibrate the brightness between $I_t(\bm{p})$ and $I_s(\bm{p})$, which can be expressed as:
\begin{align}
I_t^{\prime}(\bm{p}) = I_t(\bm{p}) + U_{\delta}(\bm{p}) \Leftrightarrow I_s(\bm{p})
\end{align}
where the $U_{\delta}(\bm{p})$ is the appearance flow produced by AFNet and $I_t^{\prime}$ is the calibrated target image. Therefore, our training network $f_{unsup}$ is formed by optimizing the calibrated photometric errors between $I_t^{\prime}$ and $I_{s \to t}$, which is defined as:
\begin{equation}
\resizebox{0.97\hsize}{!}{$
\emph{L}\left(I_{t}^{\prime}, I_{s \to t} \right) = \alpha \, \frac{(1 - \text{SSIM}(I_t^{\prime}, I_{s \to t}))}{2} + \left(1 - \alpha \right)\left\|I_t^{\prime} - I_{s \to t}\right\|_{1}
$}
\end{equation}
where $\text{SSIM}(\cdot)$ represents the structure similarity between images \cite{wang2004image} and $\alpha=0.85$ is the weight parameter. 

Using the above training process, we obtain a learning-based inverse depth estimation model that can resolve the brightness fluctuations in endoscopy. Afterwards, the estimated inverse depth can be converted to relative depth denoted as $D_{unsup}$.

\subsection{Scaling coefficient calculation with robot kinematics}

The scaling coefficient \emph{r} is calculated as follows:
\begin{equation}
\left \|{}^R\bm{t}\right\|_{2} = \emph{r} \cdot \left\|{}^I\bm{t}\right\|_{2}
\end{equation}
where ${}^R\bm{t}$ is the translation vector of the pose ${}^R\bm{T}$ calculated by robot kinematics, ${}^I\bm{t}$ denotes the translation vector of the endoscope pose ${}^I\bm{T}$ under image coordinate system, and $\left\|\cdot\right\|_{2}$ is the \text{L2} norm of a vector. However, the endoscope poses computed only from RGB images are not stable and have many noises, and the kinematics noise also makes the $^R\bm{T}$ inconsistent, which may result in scale calculation with abnormal jittering.

To solve this problem, we propose a depth-driven sliding optimization (DDSO) approach, in which the scale factor can be calculated using RGB-Depth images and robot kinematics and then optimized by a sliding window filter offline. Precisely, for each incoming brightness calibrated image $I_{i+1}^{\prime}$ and inverse depth $\emph{\text{ID}}_{i+1}$, we first convert the $\emph{\text{ID}}_{i+1}$ into relative depth $D_{i+1}$. Then, the pose of the endoscope is computed by iteratively minimizing the geometric and photometric reprojection residuals between the $\left\{ I_{i+1}^{\prime}, D_{i+1}\right\}$ and the current $\left\{I_i^{\prime}, D_i\right\}$ data. The geometric reprojection residual is expressed as:
\begin{equation}
\resizebox{0.90\hsize}{!}{$
\emph{E}_{\text{geo}} = D_{i+1}\left( W\left(\bm{p}, {}^I\bm{T}^{i+1}_i\right) \right) - |{}^I\bm{T}^{i+1}_i \cdot \pi^{-1}\left( \bm{p}, D_i(\bm{p})\right)|_z $}
\end{equation}
where ${}^I\bm{T}^{i+1}_i$ is the rigid transformation from the image at time $i$ to $i\!+\!1$, left superscript $I$ means that the pose is in image coordinate, $\bm{p}$ represents the pixel's 2D coordinates, and $|\cdot|_{z}$ is the \text{z} coordinate of a point. The function $\pi^{-1}$ denotes reprojecting 2D pixel in the image to 3D points in the camera coordinates. The warping function $W$ is defined as:
\begin{equation}
W(\bm{p}, \bm{T}) = \pi\left(\bm{T} \cdot \pi^{-1}\left(\bm{p}, D_i(\bm{p})\right)\right)
\end{equation}
where the function $\pi$ denotes projecting 3D points to 2D image coordinates.
The photometric residual is given by:
\begin{equation}
E_{\text{photo}}=I_{i+1}^{\prime}\left( W(\bm{p}, {}^I\bm{T}^{i+1}_i)\right) - I_i^{\prime}(\bm{p})
\end{equation}

After successfully obtaining the pose ${}^I\bm{T}^{i+1}_i$, we utilize the robot kinematics to calculate the relative pose ${}^R\bm{T}^{i+1}_i$ which is described under the robot base frame. Then, the scaling coefficient can be estimated by:
\begin{equation}
r_i = \frac{\left\|{}^R\bm{t}^{i+1}_i \right\|_{2}}{\left\|{}^I\bm{t}^{i+1}_i \right\|_{2}}
\end{equation}

While we can compute $r$ for each image, the measurement noise for each scale is severely large because the calculated pose and kinematics information are not accurate enough. In this case, we adopt a logarithmic moving average (LMA) with a multiplicative error model to filter the scale, which is expressed as:
\begin{equation}
\widetilde r_i = \exp \left( \frac{1}{N-1} \sum_{k=0}^{N-2} \log_{10} \left( \frac{\left\|^R\bm{t}_{i+k}^{i+k+1}\right\|_{2}}{\left\|^I\bm{t}_{i+k}^{i+k+1}\right\|_{2}} \right) \right)
\end{equation}
where $N$ is the number of images in the sliding window.

With accurate and smooth scaling coefficient generated by our DDSO method, we can scale the relative depth $D_{unsup}$.

\subsection{Robust metric depth distillation algorithm}

In this section, we adopt knowledge distillation to distill the actual scale aware ability from “teacher” into a more compact “student” network. The "student" is used for real-time metric depth estimation from monocular endoscope images without external sensors.

\begin{figure}[htb]
    \centering
    \includegraphics[width = 0.9\hsize]{"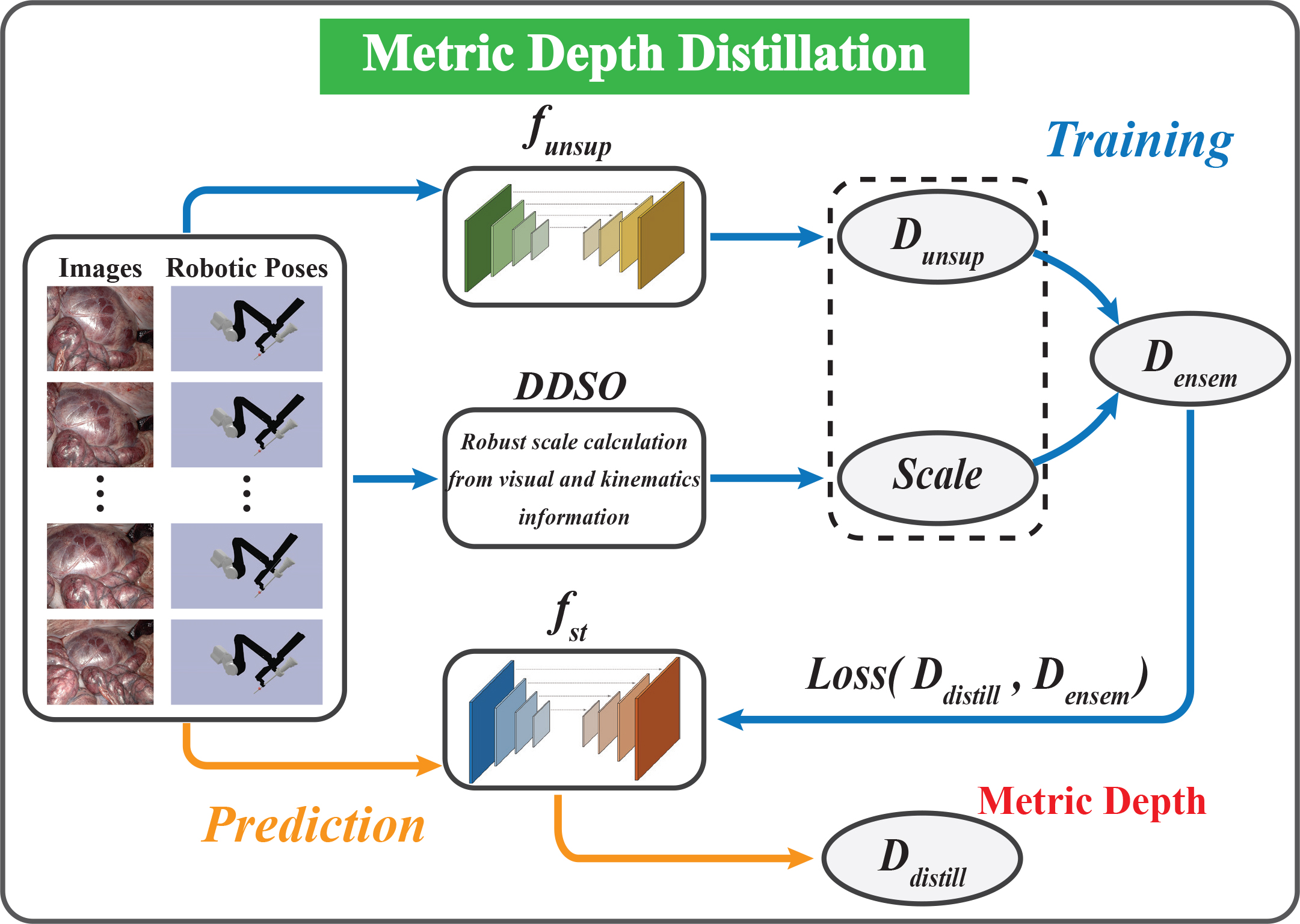"}
    \caption{Our depth distillation procedure used for metric depth estimation of monocular surgical scene. We fuse the relative depth and the computed scale into a set that represents the knowledge of the teacher. The distilled student can estimate metric and consistent depth maps as the teacher.}
    \label{depth_distill_network}
\end{figure}

To perform unsupervised distillation, as depicted in Fig. \ref{depth_distill_network}, we employ an ensemble function to combine scaling coefficient $\emph{$\widetilde r$}$ and relative depth map $D_{unsup}$ for each frame, in which each pixel value in $D_{unsup}$ is multiplied by the scale. Since $D_{unsup}$ is dense and consistent and scale $\emph{$\widetilde r$}$ is smooth, the ensemble $D_{ensem}$ is a metric and dense depth map. In our case, the "ensemble" represents embeddings extracted from monocular endoscopy and robot kinematics. To train the \emph{MetricDepthS-Net} model $f_{st}$, we build a training data that consists of $M$ monocular endoscope images and corresponding depth maps $D_{ensem}$. During the training of the model $f_{st}$, a single image is fed into the network and the corresponding metric depth $D_{distill}$ is estimated.
Then, we formulate the depth training loss $D_{\text{loss}}$ with \text{L1} and \text{SSIM}, i.e.:
\begin{equation}
\begin{split}
D_{\text{loss}} = \left( 1 - \theta\right) \cdot \left\|D_{ensem} - D_{distill}\right\|_{1} + 0.5\cdot\theta  \\
\cdot\left( 1- \text{SSIM}\left(D_{ensem},D_{distill}\right)\right)
\end{split}
\end{equation}
where the weight parameter $\theta=0.85$.

\subsection{Dense visual reconstruction of whole scene}
To provide the whole scene for the surgeons to understand the patient's anatomy, we combine monocular endoscope RGB images with the estimated metric depth $D_{distill}$ as pseudo RGB-Depth and then input them to Elasticfusion for tissue surface reconstruction \cite{whelan2016elasticfusion}. At a global level, a surfel-based tissue surface model $S_{global}$ is maintained. For every new RGB-Depth pair, the endoscope pose is computed by registering the input data to the artificial RGB-Depth image rendered from the model $S_{global}$. Then, the new depth can be integrated to $S_{global}$ after removing the outliers or artifacts. Finally, the global model $S_{global}$ will be optimized based on a space deformation approach \cite{sumner2007embedded}. The entire schematic diagram of the method is presented in Fig. \ref{overall_Framework}(b).

\section{EXPERIMENTS}

\subsection{Experiment Setup}

\textbf{Dataset:} The SCARED \cite{allan2021stereo} data is captured from porcine cadavers using a da Vinci Xi surgical robot and a AAXA P300 Neo Pico projector. Since the data consists of 9 datasets containing 4 or 5 keyframes, we call the data selected in the evaluation as $d_xk_y$ where \emph{x} and \emph{y} represent which dataset and keyframe number, respectively. 
The SCARED data provides stereo endoscopic videos with ground truth of depth maps and robot kinematics. Note that while the SCARED data owns binocular videos, we only utilize the monocular sequences in our method. The stereo images are prepared to train the comparison stereo-based approaches.

\textbf{Training:} To learn the inverse depth map estimation model, we follow the training process of AF-SfMLearner by utilizing the monocular image sequences in the SCARED. In addition, the metric depth estimation network \emph{MetricDepthS-Net} is implemented in the PyTorch and trained using Adam solver for 50 epochs with a batch size of 30. For all networks, the input images are resized to the resolution of 320\,$\times$\,256.

Extensive experiments are conducted to evaluate the performance of our framework in terms of the metric depth estimation accuracy and 3D surface reconstruction outcome. For depth evaluation, we employ the error and accuracy metrics in \cite{godard2019digging}, e.g., Root Mean Squared Error (RMSE). Besides, we adopt three metrics in \cite{ozyoruk2021endoslam}, namely, absolute trajectory error (ATE), relative translation error (RTE), and relative rotation error (RRE) to estimate the precision of camera pose in 3D reconstruction. The RMSE is used to evaluate the quantified accuracy of reconstruction results.

\begin{table}[htb]
\centering
\caption{Quantitative metric depth comparison of our results with three stereo-based methods. Best results are in boldfaced, second Best are underlined.}
\resizebox{0.98\hsize}{!}{
\begin{tabular}{cccccc}
\specialrule{0.12em}{2pt}{2pt}
\multicolumn{2}{c}{} & \multicolumn{4}{c}{\textbf{Dataset}} \\
\cline{3-6}
\specialrule{0em}{2pt}{2pt}
\multicolumn{1}{c}{} & \textbf{Methods} & $d_9k_0$ & $d_8k_2$ & $d_8k_3$ & $d_4k_3$ \\
\specialrule{0.05em}{2pt}{2pt}

\cellcolor[HTML]{F8A102}                                                & HSM \cite{yang2019hierarchical}            & \textbf{0.088}     & \textbf{0.034}     & \textbf{0.029}    & 0.284       \\
\cellcolor[HTML]{F8A102}                                                & MD\,+\,Stereo \cite{godard2019digging}     & 0.174              & \underline{0.070}  & 0.054             & 0.319        \\
\cellcolor[HTML]{F8A102}                                                & AF\_SL\,+\,Stereo \cite{shao2022self} & 0.159              & \textbf{0.034}     & \underline{0.030} & \underline{0.262}        \\

\multirow{-4}{*}{\cellcolor[HTML]{F8A102} $\text{Abs Rel} \downarrow$ }            & \textbf{\color{red}Ours Mono}       & \underline{0.129}  & 0.107              & 0.076             & \textbf{0.230}       \\
\specialrule{0.05em}{2pt}{2pt}

\cellcolor[HTML]{F8A102}                                                & HSM  \cite{yang2019hierarchical}           & \textbf{0.834}    & \textbf{0.138}    & \underline{0.120} & 7.053       \\
\cellcolor[HTML]{F8A102}                                                & MD\,+\,Stereo \cite{godard2019digging}    & 3.300             & 0.698             & 0.361             & 6.854        \\
\cellcolor[HTML]{F8A102}                                                & AF\_SL\,+\,Stereo \cite{shao2022self} & 2.419             & \underline{0.122} & \textbf{0.104}    & \underline{6.366}        \\

\multirow{-4}{*}{\cellcolor[HTML]{F8A102} $\text{Sq Rel} \downarrow$ }            & \textbf{\color{red}Ours Mono}       & \underline{1.716} & 1.043             & 0.590             & \textbf{3.256}        \\
\specialrule{0.05em}{2pt}{2pt}

\cellcolor[HTML]{F8A102}                                                & HSM    \cite{yang2019hierarchical}         & \textbf{7.694}      & \textbf{2.939}    & \underline{2.897} & 17.871      \\
\cellcolor[HTML]{F8A102}                                                & MD\,+\,Stereo  \cite{godard2019digging}   & 14.824              & 7.463             & 4.742             & \underline{16.501}      \\
\cellcolor[HTML]{F8A102}                                                & AF\_SL\,+\,Stereo \cite{shao2022self} & 13.469              & \underline{4.821} & \textbf{2.594}    & 17.563      \\

\multirow{-4}{*}{\cellcolor[HTML]{F8A102}  $\text{RMSE} \downarrow$ }             & \textbf{\color{red}Ours Mono}       & \underline{10.900}  & 6.744             & 5.974             & \textbf{11.732}      \\
\specialrule{0.05em}{2pt}{2pt}

\cellcolor[HTML]{F8A102}                                                & HSM  \cite{yang2019hierarchical}           & \textbf{0.113}  & \underline{0.046} & \underline{0.042} & \underline{0.315}     \\
\cellcolor[HTML]{F8A102}                                                & MD\,+\,Stereo  \cite{godard2019digging}   & 0.232           & 0.113             & 0.072             & 0.321     \\
\cellcolor[HTML]{F8A102}                                                & AF\_SL\,+\,Stereo \cite{shao2022self} & 0.200           & \textbf{0.041}    & \textbf{0.037}    & 0.327     \\

\multirow{-4}{*}{\cellcolor[HTML]{F8A102}  $\text{RMSE}_{log} \downarrow$}        & \textbf{\color{red}Ours Mono}     & \underline{0.155} & 0.107 & 0.087 & \textbf{0.215}     \\
\specialrule{0.05em}{2pt}{2pt}

\cellcolor[HTML]{68CBD0}                                                & HSM   \cite{yang2019hierarchical}          & \textbf{0.960}  & \underline{0.999} & \textbf{1.000}    & 0.505     \\
\cellcolor[HTML]{68CBD0}                                                & MD\,+\,Stereo \cite{godard2019digging}    & 0.696           & 0.944             & 0.981             & 0.446     \\
\cellcolor[HTML]{68CBD0}                                                & AF\_SL\,+\,Stereo \cite{shao2022self} & 0.734           & \textbf{1.000}    & \textbf{1.000}    & \underline{0.532}    \\

\multirow{-4}{*}{\cellcolor[HTML]{68CBD0}   $\delta < 1.25^1 \uparrow$}     & \textbf{\color{red}Ours Mono}       & \underline{0.865} & 0.932           & \underline{0.994} & \textbf{0.596}    \\
\specialrule{0.05em}{2pt}{2pt}

\cellcolor[HTML]{68CBD0}                                                & HSM   \cite{yang2019hierarchical}          & \textbf{0.999}  & \textbf{1.000}    & \textbf{1.000}    & \textbf{0.828}     \\
\cellcolor[HTML]{68CBD0}                                                & MD\,+\,Stereo  \cite{godard2019digging}   & 0.936           & 0.985             & \underline{0.999} & 0.782     \\
\cellcolor[HTML]{68CBD0}                                                & AF\_SL\,+\,Stereo \cite{shao2022self} & 0.987           & \textbf{1.000}    & \textbf{1.000}    & \underline{0.817}     \\

\multirow{-4}{*}{\cellcolor[HTML]{68CBD0}   $\delta < 1.25^2 \uparrow$}     & \textbf{\color{red}Ours Mono}       & \underline{0.993} & \underline{0.999} & \textbf{1.000}  & 0.782     \\
\specialrule{0.05em}{2pt}{2pt}

\cellcolor[HTML]{68CBD0}                                                & HSM     \cite{yang2019hierarchical}          & \textbf{0.999}    & \textbf{1.000}    & \textbf{1.000}   & \underline{0.969}     \\
\cellcolor[HTML]{68CBD0}                                                & MD\,+\,Stereo  \cite{godard2019digging}     & \underline{0.985} & \underline{0.995} & \textbf{1.000}   & \textbf{0.986}     \\
\cellcolor[HTML]{68CBD0}                                                & AF\_SL\,+\,Stereo \cite{shao2022self}  & \textbf{0.999}    & \textbf{1.000}    & \textbf{1.000}   & 0.945     \\

\multirow{-4}{*}{\cellcolor[HTML]{68CBD0}   $\delta < 1.25^3 \uparrow$}      & \textbf{\color{red}Ours Mono}        & \textbf{0.999}    & \textbf{1.000}    & \textbf{1.000}   & \textbf{0.986}     \\    
\specialrule{0.12em}{2pt}{2pt}

\end{tabular}
}
\label{table: depth_quantitative}
\end{table}

\subsection{Evaluation of Metric Depth Estimation} 
\textbf{Quantitative Comparison with Stereo-based Methods:} We compare the metric depth estimation accuracy of our framework with several stereo-based approaches that utilize stereo images as the training dataset, including Monodepth2 \cite{godard2019digging}, AF-SfMLearner \cite{shao2022self}, and Hierarchical Deep Stereo Matching (HSM) \cite{yang2019hierarchical}. To train the Monodepth2, we adopt the stereo pairs (the target and source images are the stereo ones) to calculate photometric loss and name the model trained by this process as MD\,+\,Stereo. For AF-SfMLearner, we add an additional loss term computed by binocular images to the original implementation for the depth estimation with accurate scale and call it AF\_SL\,+\,Stereo. Then, we use the trained model in HSM to estimate the depth of the surgical scene. The accuracy of the model has been proved in \cite{wei2021stereo}. Table \ref{table: depth_quantitative} shows the quantitative comparison outcome. Here, we randomly choose the $d_9k_0$, $d_8k_2$, $d_8k_3$, and $d_4k_3$ for evaluation, which have 903, 693, 811 and 407 frames, respectively. Our method achieves low RMSE on all four datasets, with a minimal error around 5.9 \emph{mm} and a maximal error about 11.7 \emph{mm}, demonstrating that we estimate the metric depth with high accuracy. This also indicates the consistency of the estimated depth maps on long endoscopic sequences. As can be seen in the table, our method owns a comparable metric depth estimation ability with other stereo-based methods. It is worth noted that our approach outperforms all of the compared methods on $d_4k_3$ data.

\begin{figure}[htb]
    \centering
    \includegraphics[width = 0.9\hsize]{"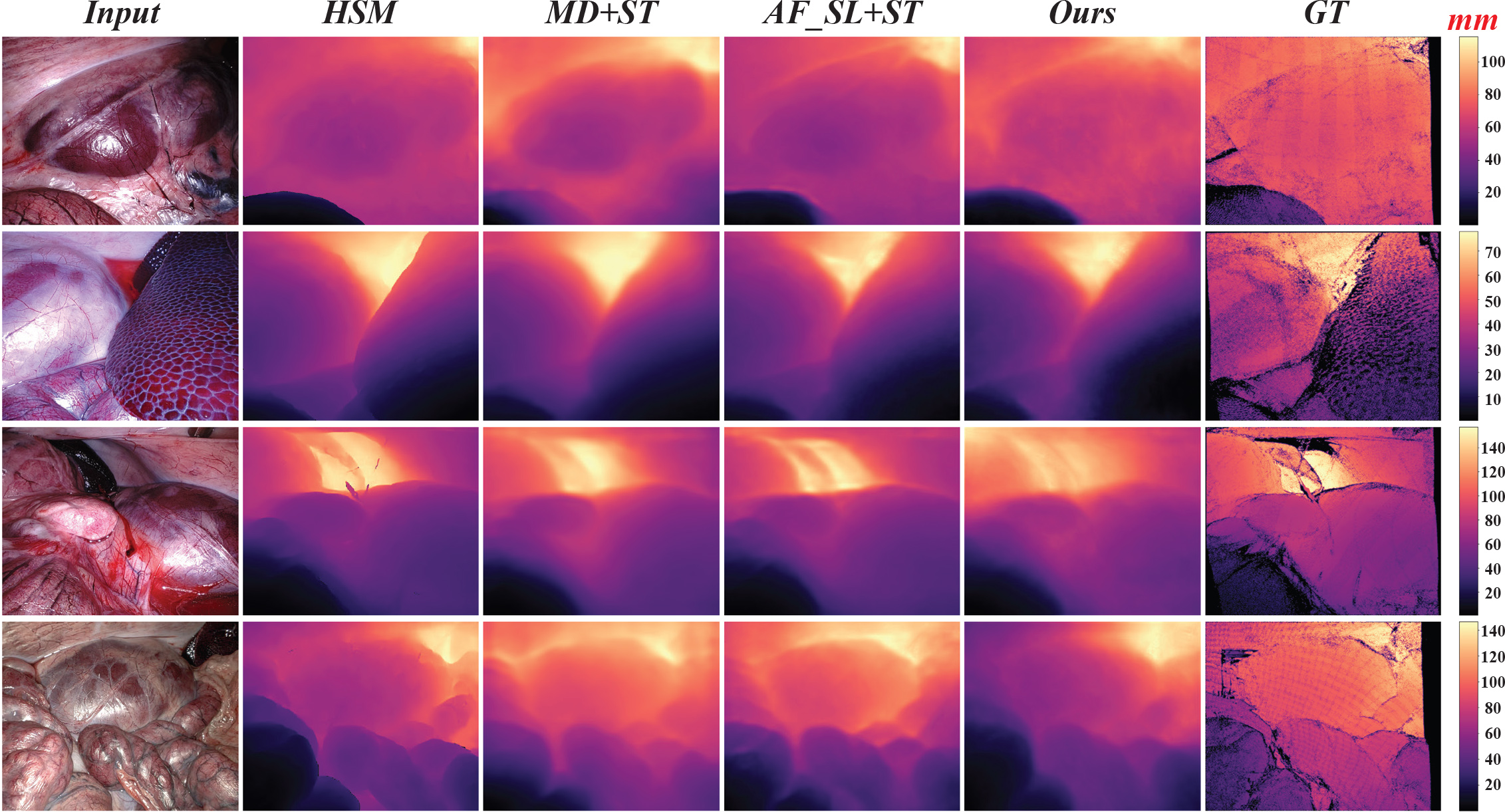"}
    \caption{Qualitative comparison with HSM \cite{yang2019hierarchical}, MD\,+\,ST \cite{godard2019digging}, AF\_SL\,+\,ST \cite{shao2022self} and ground truth (GT) on the same distance scale. The colorbar on the right represents the distance scale. Our framework can produce consistent and metric depth as stereo-based depth estimation methods. }
    \label{depth_qualitive}
\end{figure}

\textbf{Qualitative Depth Evaluation:} Four typical images from different data are chose for metric depth estimation comparison. As presented in Fig. \ref{depth_qualitive}, we put all depth maps on the same distance scale to compare our metric depth estimation method with others qualitatively. For depth results on SCARED data, the proposed framework can generate metric and smooth depth values from monocular images and perform excellent depth estimation in complex scenarios like other stereo-based methods.

\subsection{Performance of Dense Visual Reconstruction}

\begin{figure*}[htb]
    \centering
    \includegraphics[width = 0.95\hsize]{"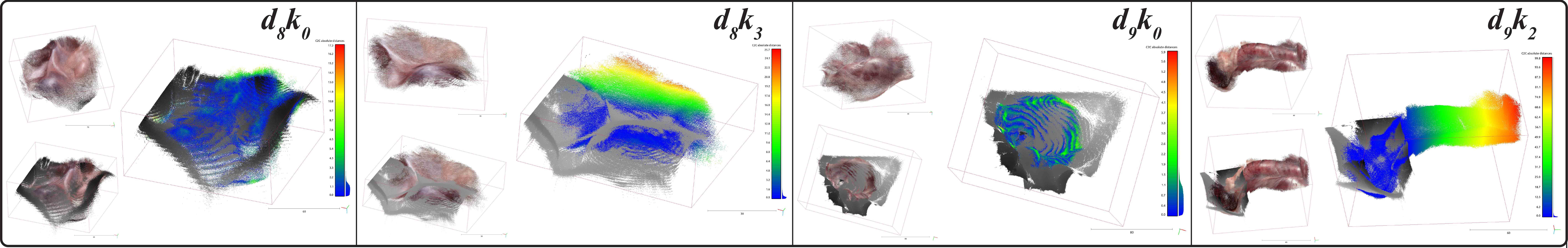"}
    \caption{Qualitative comparison on 3D reconstruction. Several typical reconstruction results are compared with the ground truth 3D structures. For each dataset, the right column represents the distance error map after registration. The left column is the reconstructed 3D tissue surface and the comparison between two structures in which the gray point cloud is the ground truth.}
    \label{3d_qualitative}
\end{figure*}

\textbf{3D Reconstruction:} We quantitatively assess the reconstruction results, and then compare the reconstructed 3D structure with ground truth 3D model on typical cases.
Since the scale of our estimated depth map is accurate, the reconstructed structure is not required to scale like other monocular scene reconstruction methods. As can be seen from Table \ref{table: 3d_quantitative}, we achieve the minimal RMSE value of 0.851 \emph{mm}, which shows the feasibility and accuracy of the dense visual reconstruction approach. The points row represents the number of points we use to calculate the RMSE in Table \ref{table: 3d_quantitative}. Due to the fusion of points and removal of outliers in reconstruction approach, the reconstruction accuracy is better than the estimated depth. Moreover, the ground truth 3D model in SCARED is only part of the whole scene compared with our reconstructed structure, so we evaluate the RMSE metric on small sets of point cloud, which also leads to low RMSE in reconstruction evaluation. Additionally, a qualitative comparison can be seen in Fig. \ref{3d_qualitative}. 
We register the reconstructed structure with the ground truth 3D point cloud and then display them using superposition. After registration, the distance error map between these two structures is calculated. Since the metric depth from \emph{MetricDepthS-Net} is gradually combined in the dense visual reconstruction, the high performance in 3D reconstruction further proves the accuracy of our metric depth estimation.

\begin{table}[htb]
\centering
\caption{Quantitative evaluation of the 3d structure}
\begin{tabular}{c||cccc}
\specialrule{0.12em}{1pt}{1pt}
Dataset & $d_8k_0$      & $d_8k_3$             & $d_9k_0$      & $d_9k_2$            \\
\specialrule{0.05em}{1pt}{1pt}
Points($10^4$)  & 42         & 48                   & 50         & 80               \\
\specialrule{0.05em}{1pt}{1pt}
RMSE(mm)     & 1.015 & 1.081   & 0.851 & 2.404 \\
\specialrule{0.12em}{1pt}{1pt}
\end{tabular}
\label{table: 3d_quantitative}
\end{table}

\begin{figure}[htb]
    \centering
    \includegraphics[width = 0.9\hsize]{"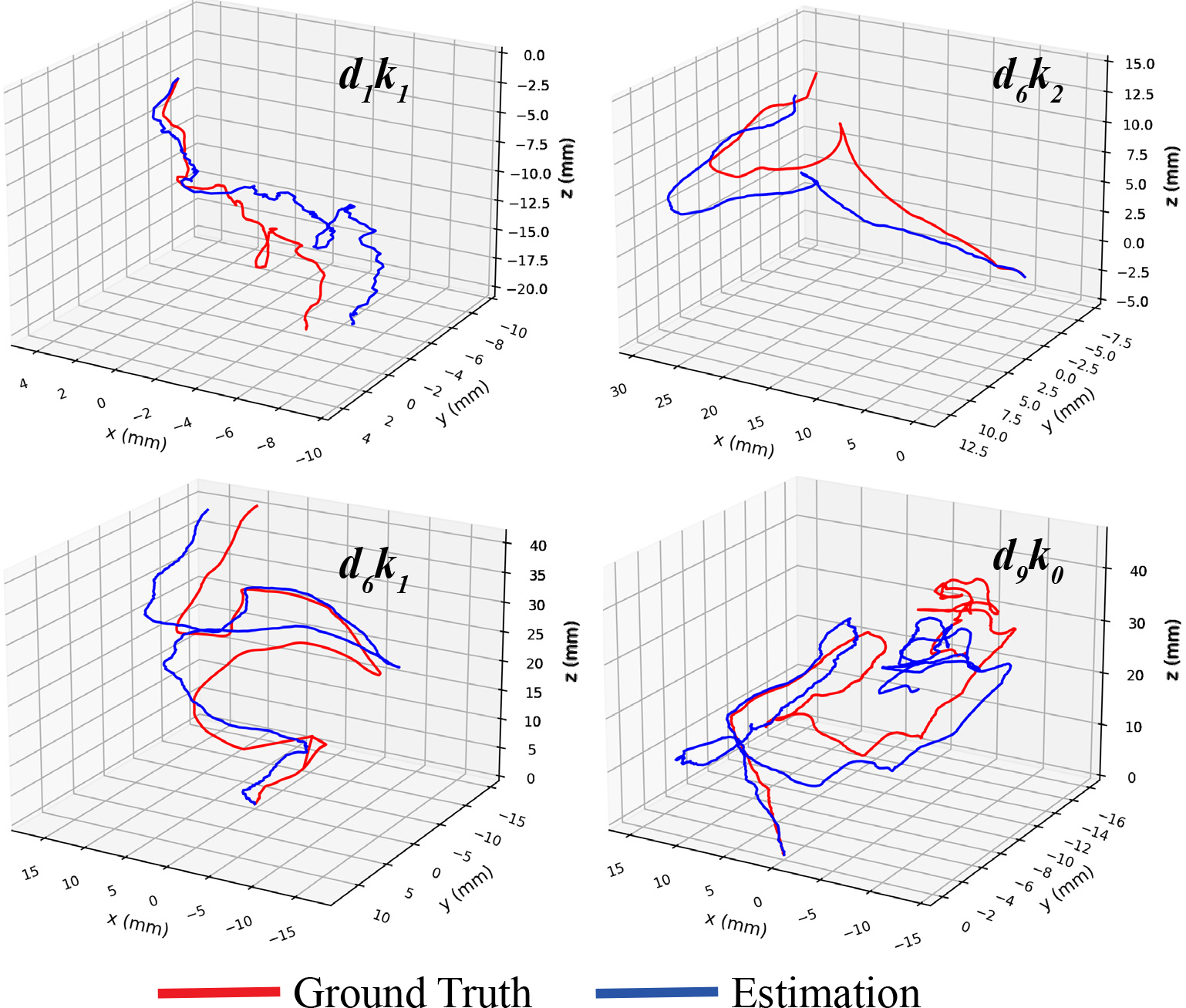"}
    \caption{Qualitative evaluation of four trajectories from the SCARED data.}
    \label{pose_qualitative}
\end{figure}

\textbf{Endoscopic Pose Estimation:} Table \ref{table: pose_quantitative} shows a quantitative evaluation of poses calculated by the dense visual reconstruction method. It is seen that the method achieves lower ATE, RTE, and RRE in pose estimation, which potentially demonstrates that the estimated metric depth maps are more accurate. In Fig. \ref{pose_qualitative},  we show the endoscope trajectories for four datasets. The result illustrates that the estimated poses are very close to the ground truth poses.

\begin{table}[htb]
\centering
\caption{Quantitative pose evaluation on four typical datasets}
\resizebox{0.98\hsize}{!}{\begin{tabular}{ccccc}
\specialrule{0.12em}{1pt}{1pt}
Dataset & $d_1k_1$      & $d_6k_2$             & $d_6k_1$      & $d_9k_0$          \\
\specialrule{0.05em}{1pt}{1pt}
Frames  & 197         & 200                   & 400         & 903                 \\
\specialrule{0.05em}{1pt}{1pt}
ATE\,[mm]     & $1.68\pm 0.97$  & $1.71\pm 0.92$    & $3.05\pm 2.44$ & $6.27\pm 3.19$  \\
\specialrule{0.05em}{1pt}{1pt}
RTE\,[mm]     & $0.10\pm 0.10$  & $0.10\pm 0.10$   & $0.33\pm 0.33$ & $0.10\pm 0.10$  \\
\specialrule{0.05em}{1pt}{1pt}
RRE\,[deg]     & $0.16\pm 0.09$  & $0.12\pm 0.09$   & $0.19\pm 0.35$ & $0.11\pm 0.07$  \\
\specialrule{0.12em}{1pt}{1pt}
\end{tabular}}
\label{table: pose_quantitative}
\end{table}

\section{CONCLUSIONS AND FUTURE WORK}
In this paper, we propose a novel framework to retrieve metric depth estimation from monocular endoscopy and then achieve dense reconstruction of complicated tissue surface. A standard monocular depth estimation network is used to obtain relative depth. Utilizing our DDSO approach, the metric scale is extracted from robot kinematics and monocular images offline. Afterwards, deep CNN is leveraged to distill the scale knowledge into a single net for metric depth estimation. Based on the metrically accurate depth, we recover the 3D structure of tissue surface by using dense visual reconstruction method. We also validate our framework quantitatively and qualitatively, indicating the efficacy and accuracy of the proposed approach in monocular metric depth estimation.

In the future, we will generate more TROS datasets to train and evaluate our framework. In addition, robot kinematics and endoscopy-related information will be directly integrated into a well-designed CNN for metric depth estimation. We then plan to develop a new optimization method for scale calculations in which only camera translation is calculated.

\bibliographystyle{IEEEtran}
\bibliography{IEEEabrv,ref}

\end{document}